\documentclass[letterpaper, 10 pt, conference]{ieeeconf}  %

\IEEEoverridecommandlockouts            %

\usepackage{amssymb}  %
\usepackage{graphicx} %
\usepackage{xcolor}
\usepackage{multirow}
\usepackage{makecell}
\usepackage{multicol}
\usepackage{booktabs}
\usepackage[utf8]{inputenc}
\usepackage{amsmath,amsfonts,booktabs,cite} %
\usepackage{siunitx,textcomp} %
\usepackage[hidelinks]{hyperref} %
\usepackage{subcaption}
\let\labelindent\relax
\usepackage[inline]{enumitem} %
\usepackage[nolist]{acronym} %
\usepackage{caption}
\usepackage{flushend}
\usepackage{tikz,standalone,pgfplots}
\pgfplotsset{compat=1.13}
\usepackage[permil]{overpic}

\newacro{bn}[BN]{Batch Normalization}
\newacro{relu}[ReLU]{Rectified Linear Unit}
\newacro{adam}[Adam]{Adaptive Moment Estimation}
\newacro{ai}[AI]{Artificial Intelligence}
\newacro{dl}[DL]{Deep Learning}
\newacro{dnn}[DNN]{Deep Neural Network}
\newacro{cnn}[CNN]{Convolutional Neural Network}
\newacro{bnn}[BNN]{Bayesian Neural Network}
\newacro{mc}[MC]{Monte-Carlo}
\newacro{dof}[DOF]{Degrees-Of-Freedom}
\newacro{ods}[USN]{Uncertain Shape Network}
\newacro{od}[SN]{Shape Network}
\newacro{gp}[GP]{Gaussian Processes}
\newacro{gmm}[GMM]{Gaussian Mixture Model}
\newacro{gmr}[GMR]{Gaussian Mixture Regression}
\newacro{em}[EM]{Expectation Maximization}
\newacro{rl}[RL]{Reinforcement Learning}

\newacro{mcmc}[MCMC]{Markov Chain Mote Carlo}
\newacro{psdf}[p-SDF]{probabilistic Signed Distance Function}
\newacro{gpis}[GPISs]{Gaussian Process Implicit Surfaces}
\newacro{va}[V]{Varley}
\newacro{spa}[SPA]{Soft Pneumatic Actuator}
\newacro{fsr}[FSR]{Force Sensitive Resistive}
\newacro{pwm}[PWM]{Pulse Width Modulation}
\newacro{rms}[RMS]{Root Mean Square}

\newacro{pi}[PI]{Proportional-Integral}
\newacro{pid}[PID]{Proportional-Integral-Derivative}
\newacro{bic}[BIC]{Bayesian Information Criterion}
\newacro{fem}[FEM]{Finite Element Method}
\newacro{ols}[OLS]{Ordinary Least Squares}

\newcommand{\figref}[1]{\hyperref[#1]{Fig.~\ref*{#1}}}
\newcommand{\tabref}[1]{\hyperref[#1]{Table~\ref*{#1}}}
\newcommand{\secref}[1]{\hyperref[#1]{Section~\ref*{#1}}}
\newcommand{\algoref}[1]{\hyperref[#1]{Algorithm~\ref*{#1}}}

\newcommand{\ra}[1]{\renewcommand{\arraystretch}{#1}}
\newcommand{\tbs}[1]{\renewcommand{\tabcolsep}{#1pt}}

\DeclareMathOperator*{\argmin}{arg\,min}

\def\methodname{Multi-FinGAN}
\def\bestcolor{(best viewed in color)}
\def\panda{\textit{Franka Emika Panda}}
\def\barrett{\textit{Barrett hand}}
\def\sota{state-of-the-art}
\def\ie{, \textit{i.e.},}
\def\eg{, \textit{e.g.},}
\def\etal{\textit{et al.}}
\def\graspit{GraspIt!}
\def\egad{EGAD!}

\def\figvspace{\vspace{-1.2em}}

\title{\LARGE \bf \methodname{}: Generative Coarse-To-Fine Sampling\\ of Multi-Finger Grasps
}

\author{Jens~Lundell, Enric~Corona, Tran~Nguyen~Le, Francesco~Verdoja, Philippe~Weinzaepfel,\\ Gr\'egory~Rogez, Francesc~Moreno-Noguer, Ville~Kyrki%
\thanks{This work was supported in part by the Academy of Finland Strategic Research Council
grant 314180, CHIST-ERA project IPALM, Finland 326304, Spain  PCI2019-103386, and by the Spanish government  with  projects  HuMoUR  TIN2017-90086-R  and  Maria de Maeztu Seal of Excellence MDM-2016-0656.}
\thanks{J.~Lundell (\texttt{jens.lundell{@}aalto.fi}), T.~Nguyen~Le, F.~Verdoja, and V.~Kyrki are with School of Electrical Engineering, Aalto University, Finland. E.~Corona and F.~Moreno-Noguer are with Institut de Robòtica i Informàtica Industrial, CSIC-UPC, Barcelona, Spain. G.~Rogez and P.~Weinzaepfel are with NAVER LABS Europe, Meylan, France.}}

\makeatletter
\let\@oldmaketitle\@maketitle%
\renewcommand{\@maketitle}{\@oldmaketitle%
  \includegraphics[width=\linewidth]{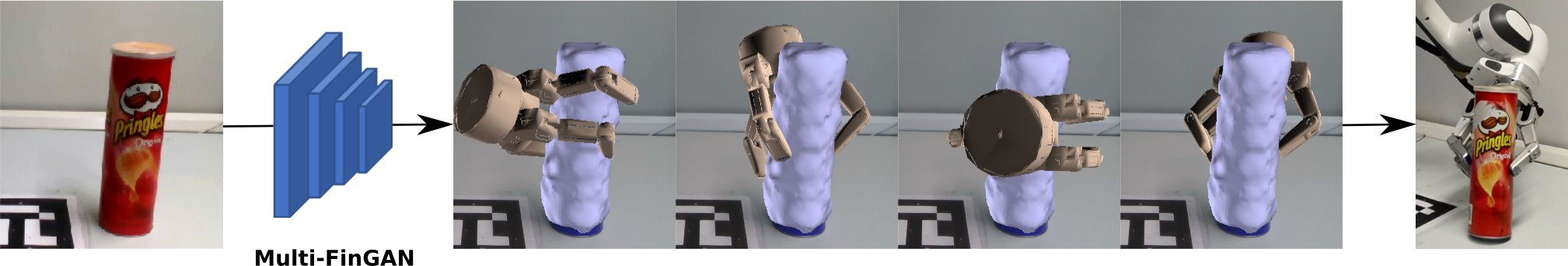}
  \captionof{figure}{\label{fig:real_experiment}From an input RBG-D image, \methodname{} generates a diverse set of grasps from all around the object in about a second, and then executes the highest scoring grasp on the real robot. \figvspace{}
}}%
\makeatother

\begin{document}

\maketitle

\thispagestyle{empty}
\pagestyle{empty}

\begin{abstract}
While there exists many methods for manipulating rigid objects with parallel-jaw grippers, grasping with multi-finger robotic hands remains a quite unexplored research topic. Reasoning and planning collision-free trajectories on the additional degrees of freedom of several fingers represents an important challenge that, so far, involves computationally costly and slow processes.
In this work, we present \emph{\methodname{}}, a fast generative multi-finger grasp sampling method that synthesizes high quality grasps directly from RGB-D images in about a second. We achieve this by training in an end-to-end fashion a coarse-to-fine model composed of a classification network that distinguishes grasp types according to a specific taxonomy and a refinement network that produces refined grasp poses and joint angles. We experimentally validate and benchmark our method against a standard grasp-sampling method on 790 grasps in simulation and 20 grasps on a real \panda{}. All experimental results using our method show consistent improvements both in terms of grasp quality metrics and grasp success rate. Remarkably, our approach is up to 20-30 times faster than the baseline, a significant improvement that opens the door to feedback-based grasp re-planning and task informative grasping. Code is available at \url{https://irobotics.aalto.fi/multi-fingan/}.
\end{abstract}

\section{Introduction}
\label{sec:introduction}

Generating multi-fingered grasps for unknown objects such as the one shown in \figref{fig:real_experiment} is still non-trivial and considerably more challenging than using parallel-jaw grippers for the same task. However, by actuating more joints, multi-fingered grippers allow a robot to perform more advanced manipulations, including precision grasping flat disks or power grasping spherical objects.

Typical methods for multi-fingered grasp generation require known 3D object models and poses in order to sample a large space of candidate grasps and then evaluate them based on physical grasping metrics such as the $\epsilon$-quality~\cite{ferrari1992planning}. In case of unknown object poses, pose estimation is the {\em de facto} standard solution~\cite{kleeberger2020survey}, while for unknown models estimating the shape through\eg{} shape completion or mirroring has shown to work well in many scenarios~\cite{varley_shape_2017,lundell2019robust,bohg2011mind}. Nevertheless, generating good candidate grasps with these methods is computationally expensive as it usually relies on a stochastic search process such as simulated annealing over a large search space. For instance, the search space for the robotic hand we consider in this work, the \barrett{} seen in \figref{fig:real_experiment}, has 7~\ac{dof} which together with the 6D object poses (3 rotations and 3 translations) result in a 13 dimensional search space. Despite clever solutions to reduce the search space, such as limiting the search over eigengrasps~\cite{ciocarlie2007dexterous}, the process is still inherently slow (in the order of several tens of seconds) due to the stochastic search procedure.

In this work, we present a deep network inspired by recent work from the computer vision community on human hand grasp synthesis~\cite{corona2020ganhand,hasson2019learning} that can generate and evaluate multi-finger grasps on unknown objects in roughly a second. To achieve this, we devise a generative architecture for coarse-to-fine grasp sampling named \emph{\methodname{}} that is purely trained on synthetic data. Especially the integration of a novel parameter-free finger refinement layer based on a fully differentiable forward kinematics layer of the \barrett{} facilitates fast learning and robust grasp generation.

The proposed sampling method is quantitatively evaluated in both simulation, where we compare over 790 grasps against the baseline in terms of analytical grasp quality metrics, and on a real \panda{} equipped with a \barrett, where we evaluate grasp success rate on 10 grasps per method. In both cases, our approach demonstrates a significant reduction in running time compared to the baseline while still generating grasps with high quality metrics and success rate.

In summary, the  main  contributions  of  this  work  are:
\begin{enumerate*}[label=(\roman*)]
	\item a novel generative method for multi-finger grasp selection that enables fast sampling with high coverage;
	\item a novel loss function for guiding grasps towards the object while minimizing interpenetration;
	and
	\item an empirical evaluation of the proposed method against \sota{}, presenting, both in simulation and on real hardware, improvements in terms of running time, grasp ranking, and grasp success rate.
\end{enumerate*}

\section{Related work}
\label{sec:related_work}

When considering parallel-jaw grippers, a large corpus of data-driven generative~\cite{morrison2018closing,mousavian20196,murali20206,pinto2017asymmetric,viereck2017learning} and classification based~\cite{mahler2017dex,satish2019policy} grasping methods exist. Many of these approaches~\cite{morrison2018closing,mahler2017dex} reach a grasp success rate over 90\% on a wide variety of objects by constraining to top-down-only grasps. However, as recently discussed by Wu~\etal{}~\cite{wu2020generative}, the simplifications made in these works exclude many solutions that could be used for applications like semantic and affordance grasping \cite{kokic2017affordance} or multi-finger grasping for dexterous manipulation.

Despite the limitations of parallel-jaw grippers, alternative methods using multi-fingered hands have not seen as much development~\cite{varley2015generating,lu2020multi,wu2020generative,aktas2019deep,lu2020active,liu2019generating,kokic2017affordance}. These approaches generally under-perform both in terms of running time and grasp success rate compared to their parallel-jaw counterparts.

One of the earliest deep-learning-based multi-finger grasping work trained a network to detect the palm and fingertip positions of stable grasps directly from an RGB-D view~\cite{varley2015generating}. That method achieved a 75\% grasp success rate on 8 objects but relied on an external planner (\graspit{}) at run-time to generate grasp samples which made the method slow (16.6 seconds on average to generate a grasp). Our generative method also takes RGB-D images as input but does not require any external planner, making it a much faster solution.

To remove the need of a slow external planner, recent work also focused on generating grasps~\cite{liu2019generating,aktas2019deep,lu2020multi}. For instance, in~\cite{liu2019generating},  the authors train a network that regresses from a voxel grid representation of the object to the output pose and configuration of the gripper. Similar to ours, this work also employs the known forward kinematics equation of the gripper to compute a collision loss. At run-time the generated grasp is refined by searching over all ground-truth grasps, selecting the grasp closest to the generated one. The main drawback is that the grasps are viewpoint dependent so, to generate grasps in all possible locations around the object, the input representation needs to be rotated. Our method, on the other hand, can generate grasps from any orientation in just a single forward pass.

Aktas \etal{} \cite{aktas2019deep} proposed a generative-evaluative model which both generates grasps and subsequently test them. Grasps are produced through stochastic hill-climbing on a product of experts, which is a sequential and time-consuming process. Our model, on the other hand, only requires one forward pass to generate a grasp that is then evaluated by computing analytical quality metrics~\cite{ferrari1992planning}.

The work most similar to ours is by Lu~\etal{}~\cite{lu2020multi}. It proposes a deep network that, given an initial grasp configuration and a RGB-D or voxel representation of the object, optimizes a hand pose and finger joints to increase grasp success. The proposed method reached an average grasp success rate of 57.5\% but requires roughly 5--10 seconds to generate a grasp. That work was later improved in terms of data-efficiency~\cite{lu2020active}, but still the method required 3--10 seconds to generate a grasp. Our work, in comparison, does not require an explicit initial hand configuration as the network implicitly learns to predict such a configuration. Moreover, our method is much faster at generating grasps.

Tangential to training a multi-finger grasp sampler with supervised learning is to use \ac{rl}. In~\cite{wu2020generative}, Wu~\etal{} learned a deep 6-\ac{dof} multi-finger grasping policy directly from RGB-D inputs. That work introduced a novel attention mechanism that zooms in and focuses on sub-regions of the depth image to achieve better grasps in dense clutter. Although the policy was trained purely in simulation, it transferred seamlessly to the real world and attained a high grasp success rate on a diverse set of objects. Nevertheless, training such a method requires an elaborate simulation setup and fine-tuning of hyper-parameters for the \ac{rl} method to work well.

\section{Problem formulation}
\label{sec:prob_form}

In this work, we consider the problem of grasping unknown objects with a multi-fingered robotic hand. This implies producing a grasp that does not interpenetrate the object but has several contact points with it. More formally, we train a model $\mathcal{M}$ that takes as input an RGB-D image $\mathbf{I}$ and produces a grasp type $c$, a 6D gripper pose $\mathbf{p}$, and a valid hand joint configuration $\mathbf{q}$:
\begin{equation*}
    \mathcal{M}: \mathbf{I} \implies \left\{c, \mathbf{p}, \mathbf{q}\right\}\enspace.
\end{equation*}

We represent the hand joint configuration $\mathbf{q}$ by a 7-\ac{dof} \barrett{} shown in \figref{fig:real_experiment}. $c$ is a coarse grasp class within the 33-grasp taxonomy listed in~\cite{feix2015grasp}.  Due to physical constraints, the \barrett{} can achieve only 7 out of these 33 grasp types, namely: small wrap, medium wrap, large wrap, power sphere, precision sphere, precision grasp, and pinch grasp. All grasps are in the object's center of reference.

Furthermore, we assume that the hand joint configuration $\mathbf{q}$ leaves a small gap between the fingers and the object as the gripper executes a close-hand primitive before attempting to actually lift the object. This assumption is reasonable to limit the impact of sensing uncertainties in the real world as we avoid the need to generate precise configurations that actually touch the surface of the object.

\section{Method}
\label{sec:method}

\begin{figure*}
	\centering
	\begin{overpic}[width=\linewidth]{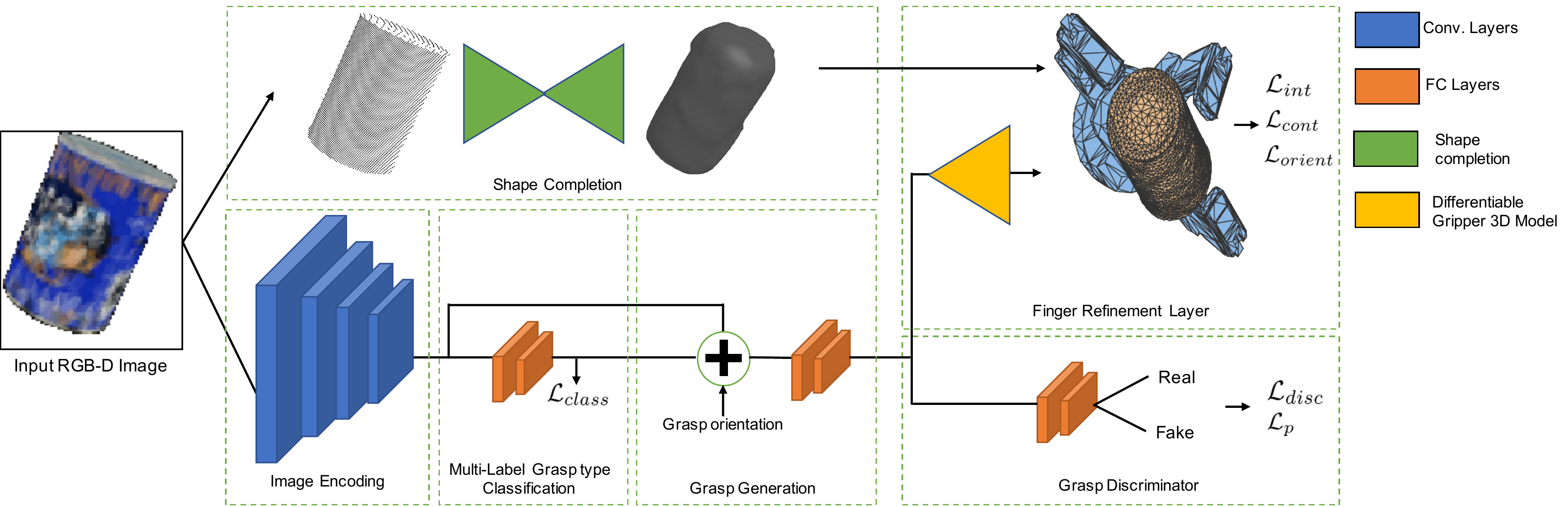}%
	\end{overpic}
	\caption{\label{fig:network}The architecture of \methodname{}.}
 \figvspace{}
\end{figure*}
 
Our model for generating 6D multi-finger grasps inspired from~\cite{corona2020ganhand} is visualized in \figref{fig:network}. It consists of 6 different submodules: Shape Completion, Image Encoding, Multi-Label Grasp type Classification,  Grasp Generation, Discriminator and Finger Refinement. All these modules are novel for robotic grasping except ``shape completion'', which was used in~\cite{lundell2020beyond} to shape complete voxelized input point-clouds using a deep network, and ``image encoding'', which is a pre-trained ResNet-50\cite{he2016deep}. In the following subsections we will present each of the novel modules and their function. Finally, we will present the loss functions used to train the complete generative grasp planning architecture end-to-end.

\subsection{Multi-Label Grasp Type Classification}

The task of the Multi-Label Grasp Type Classification network is to classify which of the seven predefined grasp classes are feasible for a given object. For this purpose, the classification network is fed with the image representation of an object $\textrm{E}(\mathbf{I})$, where $\textrm{E}(\cdot)$ is a ResNet-50 encoding, and produces as output a grasp type $c$.

As objects can often be associated with multiple correct grasp categories, we frame the problem as a multi-label classification task. As such, we use a Sigmoid activation function at the output and the binary cross-entropy loss $\mathcal{L}_{class}$ to train the network. To later choose one grasp among all the possible ones, we threshold the output to 0.5 and randomly choose one grasp type $c$ that is classified as valid for that object.

\subsection{Grasp Generation}

The grasp type $c$ estimated by the classification network is associated with a coarse hand configuration $\mathbf{q}_c$ \ie{} the average joint angles. The task of the Grasp Generation is to generate a first refinement of the hand configuration $\mathbf{q}_r = \mathbf{q}_c + \Delta \mathbf{q}$ along with a 6D hand pose $\mathbf{p} = \left\{\mathbf{t},~\mathbf{r}\right\}$  where $\mathbf{t}$ is a translation and $\mathbf{r}$ is a normalized axis-angle rotation. %

Given the object mesh, we estimate the center of it $\mathbf{t}_{0}$ and have the network refine this translation as $\mathbf{t}^* = \mathbf{t}_{0}+ \Delta \mathbf{t}$. Similarly, we represent the refinement of the hand's rotation as $\mathbf{r}^* = \mathbf{r}_0 + \Delta \mathbf{r}$. At training time the input rotation $\mathbf{r}_0$ is set to a rotation of a ground-truth grasp with added zero-mean Gaussian noise while at test-time we sample uniform rotations and feed these to the network. The center of the object mesh $\mathbf{t}_{0}$ is known when training and during the simulation experiments, while during the physical experiments we use the center of the shape completed object.

All in all, the network is represented as a fully-connected residual network which takes as input the initial hand configuration, the object center, a noisy rotation, and the image encoding $\left\{\mathbf{q}_c, \mathbf{t}_{0}, \mathbf{r}_0, \textrm{E}(\mathbf{I})\right\}$ and produces individual refinements $\left\{\Delta \mathbf{q}, \Delta \mathbf{t}, \Delta \mathbf{r}\right\}$.

\subsection{Finger Refinement Layer}

The Finger Refinement Layer is responsible to further refine the hand representation $\mathbf{q}_r$. To this end, we propose a novel fully-differentiable and parameter-free layer based on the forward kinematics equation of the \barrett{}. This layer takes as input the pose of the gripper $\mathbf{p}$ and the coarse gripper representation $\mathbf{q}_r$ and produces an optimized gripper configuration $\mathbf{q}^*=\mathbf{q}_r+\Delta \mathbf{q}^*$ that is close to the surface of the object but not in collision with it. 

We optimize each finger independently with respect to the estimated object mesh.
We denote $\Delta \mathbf{q_j}^*$ as the optimized position of joint $j$.
This is calculated by rotating the articulated finger within its predefined physical limit $\theta_j$ until the distance $\delta_{\theta_j}$ between the finger vertices ${V_i^{\theta_j}}$ and the object vertices ${O_k}$ implies a contact between finger and object. Hand-Object contact is parameterized by a threshold hyperparameter $t_d$, following
\begin{align}\label{eq:delta_q}
    \begin{split}
        \Delta \mathbf{q}^*_j &= \argmin_{\theta_j} \{\delta_{\theta_j}+\epsilon-\mathbf{q}_{r,j}\}\hspace{0.3cm} \forall \theta_j \; s.t. \; {\delta_{\theta_j} < t_d}\enspace,\\
        \delta_{\theta_j} &= \min_{i}(\min_{k}(\|V_i^{\theta_j},O_k\|))\enspace.
    \end{split}
\end{align}

Note that we could have simply set $\mathbf{q}^* = \argmin_{\theta} \delta_\theta$ for each joint as was proposed in~\cite{corona2020ganhand} for a human hand model. However this would break backward differentiability, and instead, we explicitly calculate $\Delta \mathbf{q}^*$ and add it to $\mathbf{q}_r$. 

The \barrett{} consists of three fingers made of two links (a proximal and a distal one). As such, \eqref{eq:delta_q} is solved for $j=1,\dots,6$, where we first rotate the proximal joints until contact and then proceed with the distal joints. To avoid interpenetration between the object and the gripper, we add an offset $\epsilon$ which we heuristically set to 0.5 cm in  our experiments.

\subsection{Discriminator network}

Since the network does not have any supervision except for the classification task, we need to enforce that the generated grasps are realistic. To this end, we add a Wasserstein discriminator D~\cite{martin2017wasserstein} and train it with the gradient penalty~\cite{gulrajani2017improved}. More specifically, the objective to minimize using the grasp generation module G is
\begin{align}
\begin{split}
    \mathcal{L}_{disc} =& \mathop{\mathbb{E}}\left[\text{D}(\text{G}(\textrm{E}(\mathbf{I}),\mathbf{q}_c,\mathbf{t}_0,\mathbf{r}_0)\right]-\mathop{\mathbb{E}}\left[\text{D}(\widehat{\mathbf{q}},\widehat{\mathbf{t}},\widehat{\mathbf{r}})\right]\enspace,\\
    \mathcal{L}_{gp} =& \mathop{\mathbb{E}}\left[\left(\|\nabla_{\widetilde{\mathbf{q}},\widetilde{\mathbf{T}},\widetilde{\mathbf{r}}} \text{D}(\widetilde{\mathbf{q}},\widetilde{\mathbf{t}},\widetilde{\mathbf{r}}) \|_2 -1\right)^2\right]\enspace,
\end{split}
\end{align}
where $\widehat{\mathbf{q}}$, $\widehat{\mathbf{t}}$, and $\widehat{\mathbf{r}}$ are samples from the ground-truth data and $\widetilde{\mathbf{q}}$, $\widetilde{\mathbf{t}}$, and $\widetilde{\mathbf{r}}$ are linear interpolations between predictions and those ground-truth samples.

\subsection{Complementary loss functions}

While the discriminator loss $\mathcal{L}_{disc}$ helps in producing realistic looking grasps, it alone is not sufficient to guide the learning problem enough. Therefore, we propose a set of complementary losses.

To ensure that the generated grasps are close to the object, we add a contact loss
\begin{align}
\mathcal{L}_{cont} = \frac{1}{|V_{cont}|}\sum_{v\in V_{cont}}\min_{k}\|v,O_k\|_2\enspace,   
\end{align}
where $O_k$ are the object vertices and $V_{cont}$ are vertices on the hand that are often in contact with the object in our ground-truth grasps. We calculate $V_{cont}$ as the vertices that are closer than 5 mm to the object in at least 8\% of the ground-truth grasps. These are mainly located on the finger tips and the palm of the hand. 

For a grasp to be successful, the gripper should be rotated towards the object of interest. To promote such behaviour, we add a loss function that penalizes the gripper if its approach direction $\mathbf{\hat{a}}$ is pointing away from the vector $\mathbf{\hat{o}}$ connecting the hand to the object's center:
\begin{align}
\mathcal{L}_{orient} = 1 - \mathbf{\hat{a}}^\top\mathbf{\hat{o}}\enspace.
\end{align}

Finally, for a grasp to be successful it cannot interpenetrate the object. To encourage such behaviour, we add a loss that penalizes the distance between vertices $V_i$ that are inside the object and the closest object vertex 
\begin{align}
\mathcal{L}_{int} = \frac{1}{|V_i|}\sum_{\mathbf{v}\in V_i} \text{A}_\mathbf{v}\min_k \|\mathbf{v},O_k\|_2\enspace,
\end{align}
where $\text{A}_{\mathbf{v}}$ is the average area of the incident faces of the vertex that is inside the object. Since uniform mesh tessellation cannot usually be assumed in robotics\eg{} \figref{fig:grasp_in_simulation}, we add the term $\text{A}_{\mathbf{v}}$ to be robust to non-uniform tessellation.

Finally, the total loss is a linear combination of all the separate loss functions $\mathcal{L}_{tot}=w_{class}\mathcal{L}_{class}+w_{disc}\mathcal{L}_{disc}+w_{gp}\mathcal{L}_{gp}+w_{cont}\mathcal{L}_{cont}+w_{int}\mathcal{L}_{int}+w_{orient}\mathcal{L}_{orient}$, where each individual loss contribution is given a corresponding weight. The model is trained end-to-end.

\subsection{Implementation details}

The network was implemented in PyTorch 1.5.1. We use a pre-trained ResNet-50 \cite{he2016deep} as the image encoder. The model was trained on 30 objects from the YCB object set and for each of them we synthetically rendered 100 novel viewpoints, some examples are shown in \figref{fig:synthetic_training_data}. The images are resized to 256x256. We trained our networks with a learning rate of $1\cdot10^{-4}$ and a batch size of 100. The weights of the loss functions were experimentally set to $w_{class}=1$, $ w_{disc}=1$, $w_{gp}=10$, $w_{cont}=100$, $w_{int}=100$, $w_{orient}=1$. The generator is trained once every 5 forward passes to improve the relative quality of the discriminator. We trained the networks for 800 epochs where we linearly reduced the learning rate for the last 400 epochs.

\section{{Experiments and Results}}
\label{sec:exp_and_res}

The three main questions we wanted to answer in the experiments were:
\begin{enumerate}
    \item Is \methodname{} able to generate high quality grasps?
    \item What are the contributions of the proposed loss functions?
    \item Is our generative grasp sampler, which is purely trained on synthetic data, able to transfer to real objects?
\end{enumerate}

In order to provide justified answers to these questions, we conducted three separate experiments. In the first experiment (\secref{exp:grasp_in_sim}) we evaluate grasp quality and hand-object interpenetration in simulation. In the second experiment (\secref{sec:ablation_study}) we do an ablation study over the proposed loss functions and in  (\secref{exp:sim_to_real}) we finally evaluate grasp success rate on real hardware.

\subsection{Dataset}
\label{sec:exp_setup}
\begin{figure}%
	\centering
	\begin{subfigure}[b]{.69\linewidth}
		\centering
		\includegraphics[width=0.3\linewidth]{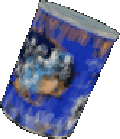}%
		\includegraphics[width=0.3\linewidth]{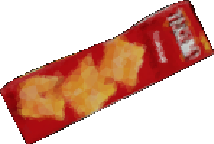}%
		\includegraphics[width=0.3\linewidth]{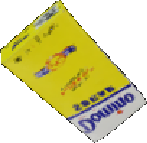}%
		\caption{\label{fig:synthetic_training_data}}
	\end{subfigure}
	\begin{subfigure}[b]{.29\linewidth}
		\centering
		\includegraphics[width=\linewidth]{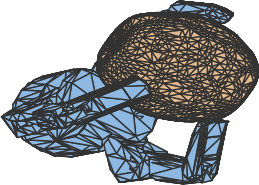}%
	    \caption{\label{fig:grasp_in_simulation}}
	\end{subfigure}
	\caption{\label{fig:examples}Example of three synthetic RGB images used for training (a), and a grasp generated by our method (b).}
    \figvspace{}
\end{figure}

To train our model, we manually generate a dataset of grasps on the YCB objects~\cite{calli2015ycb} using a \barrett{} in \graspit{}~\cite{miller2004graspit}. As previously mentioned, the hand can only attain 7 of the 33 grasp types listed in~\cite{feix2015grasp} and we therefore categorize each grasp according to these. As a final step, we generate additional grasps around the symmetry axes of the objects. In total, this amounts to over 4000 labeled grasps. \label{It seems a bit odd that the data-set is introduced here while the implementation details are introduced earlier.}

\subsection{Grasping in Simulation}
\label{exp:grasp_in_sim}

In the simulated grasping experiment, we evaluate how good our method is at producing high quality grasps that are not interpenetrating the object. We test our model on two different datasets: 33 objects from the YCB object set~\cite{calli2015ycb} and 49 objects from the  recent \egad{} dataset~\cite{morrison2020egad}. The YCB object set contains both object models we trained on, and models that were held out during training; the \egad{} dataset contains completely novel objects. 

We benchmark against the simulated annealing planner in \graspit{}~\cite{ciocarlie2007dimensionality} that ran for 75000 steps to generate 360 grasp candidates on average. To evaluate the quality of a grasp, we used the $\epsilon$-quality metric which represents the radius of the largest 6D ball centered at the origin that can be enclosed by the convex hull of the wrench space~\cite{miller1999examples}.
With our method, we render 5 different viewpoints for each object and 110 grasps per viewpoint. Out of this pool of grasps, we report the average performance on the 10  top-scoring grasps according to the $\epsilon$-metric. We also average the performance of the 10 top-scoring grasps found with the baseline method.  In total this amounts to 790 grasps per method. 

\begin{table}
    \centering
    \ra{1.3}\tbs{7}
    \caption{\label{tb:sim_exp_summary}Simulation experiment results. $\uparrow$: higher the better; $\downarrow$: lower the better.}
    \begin{tabular}{@{}lcccc@{}}
        \toprule
        & \multicolumn{2}{c}{\graspit} & \multicolumn{2}{c}{\methodname{}} \\
        \cmidrule(lr){2-3} \cmidrule(lr){4-5} & YCB & \egad{} & YCB & \egad{}\\
        \midrule
        $\epsilon$-quality $\uparrow$                  & 0.75  & 0.75  & \textbf{0.85} & \textbf{0.86}\\
        Interpenetration ($\text{cm}^3$) $\downarrow$  & \textbf{2.63}  & 1.03  & 6.88 & \textbf{1.01} \\
        Grasp Sampling (sec.) $\downarrow$             & 32.79 & 29.88 & \textbf{1.33} & \textbf{1.28} \\
        \bottomrule
    \end{tabular}
 \figvspace{}
\end{table}

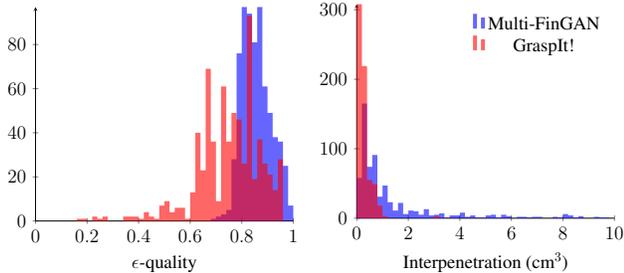
\begin{figure}
    \centering
        \pgfplotsset{ every non boxed x axis/.append style={x axis line style=-} }
    \begin{tikzpicture}[scale=0.5]
        \Large
        \begin{axis}[
            ybar,
            bar width=0.02,
            bar shift=0pt,
            xlabel=$\epsilon$-quality,
            axis x line=bottom,
            axis y line=left,
            xtick distance=0.2,
            xmin=0,xmax=1
        ]
            \addplot[fill=blue,fill opacity=0.6,draw opacity=0] table [x=x, y=ours] {figures/quality.csv};
            \addplot[fill=red,fill opacity=0.6,draw opacity=0] table [x=x, y=graspit] {figures/quality.csv};
        \end{axis}
    \end{tikzpicture}
    \begin{tikzpicture}[scale=0.5]
        \Large
        \begin{axis}[
            ybar,
            bar width=0.2,
            bar shift=0pt,
            legend pos=north east,
            legend style={draw=none},
            xlabel=Interpenetration ($\text{cm}^3$),
            axis x line=bottom,
            axis y line=left,
            xtick distance=2,
            xmin=0,xmax=10
        ]
            \addplot[fill=blue,fill opacity=0.6,draw opacity=0] table [x=x, y=ours] {figures/inter.csv};
            \addplot[fill=red,fill opacity=0.6,draw opacity=0] table [x=x, y=graspit] {figures/inter.csv};
            \legend{\methodname{}, \graspit{}}
        \end{axis}
    \end{tikzpicture}
    \caption{\label{fig:plots} Histograms showing all results obtained on both datasets by our approach and the baseline in terms of $\epsilon$-quality and interpenetration \bestcolor{}.}
    \figvspace{}
\end{figure}

\tabref{tb:sim_exp_summary} and \figref{fig:plots} show the simulation results and \figref{fig:grasp_in_simulation} an example grasp using our method. To analyze the statistical differences between the methods we used a one sided Wilcoxon signed-rank test. Based on these results we can draw several interesting conclusions. \methodname{} is able to achieve statistically better results ($\alpha = 0.001$) in terms of quality. The histogram in \figref{fig:plots} is also confirming this result, showing that our data-driven grasp planning method is more consistent than the baseline at generating high quality grasps. However, in terms of interpenetration, \graspit{} shows a statistically significant improvement over our method ($\alpha = 0.001$). This result is most likely due to our method reaching a higher interpenetration on the YCB objects because of the presence of large objects which were not in the training-set. \egad{}, on the other hand, contains objects scaled to the size of the hand and our method achieves a low interpenetration on those. One thing to note, though, is that when looking at the interpenetration histogram in \figref{fig:plots}, the two methods do not perform radically differently. Finally, our method generates, on average, a grasp in around a second compared to the 30 seconds required by \graspit{}, which makes \methodname{} 30 times faster than the baseline; a difference which is once again statistically significant ($\alpha = 0.001$).

\subsection{Ablation study}
\label{sec:ablation_study}
\begin{table}
    \centering
	\ra{1.3}\tbs{7}
	\caption{\label{tb:ablation_study}Ablation study on \egad{}}
    \begin{tabular}{@{}lrrrrr@{}}
	    \toprule
        Loss removed & none & $\mathcal{L}_{int}$ & $\mathcal{L}_{cont}$ & $\mathcal{L}_{orient}$ & $\mathcal{L}_{disc}$ \\
        \midrule
        $\epsilon$-quality $\uparrow$                 & \textbf{0.86} & 0.77  & 0.83 & 0.85 & 0.74 \\
        Interpenetr. ($\text{cm}^3$) $\downarrow$ & \textbf{1.01} & 56.24  & 3.63 & 2.99  & 27.50 \\
        \bottomrule
    \end{tabular}
 \figvspace{}
\end{table}

\begin{figure}
	\centering
	\includegraphics[width=.8\linewidth]{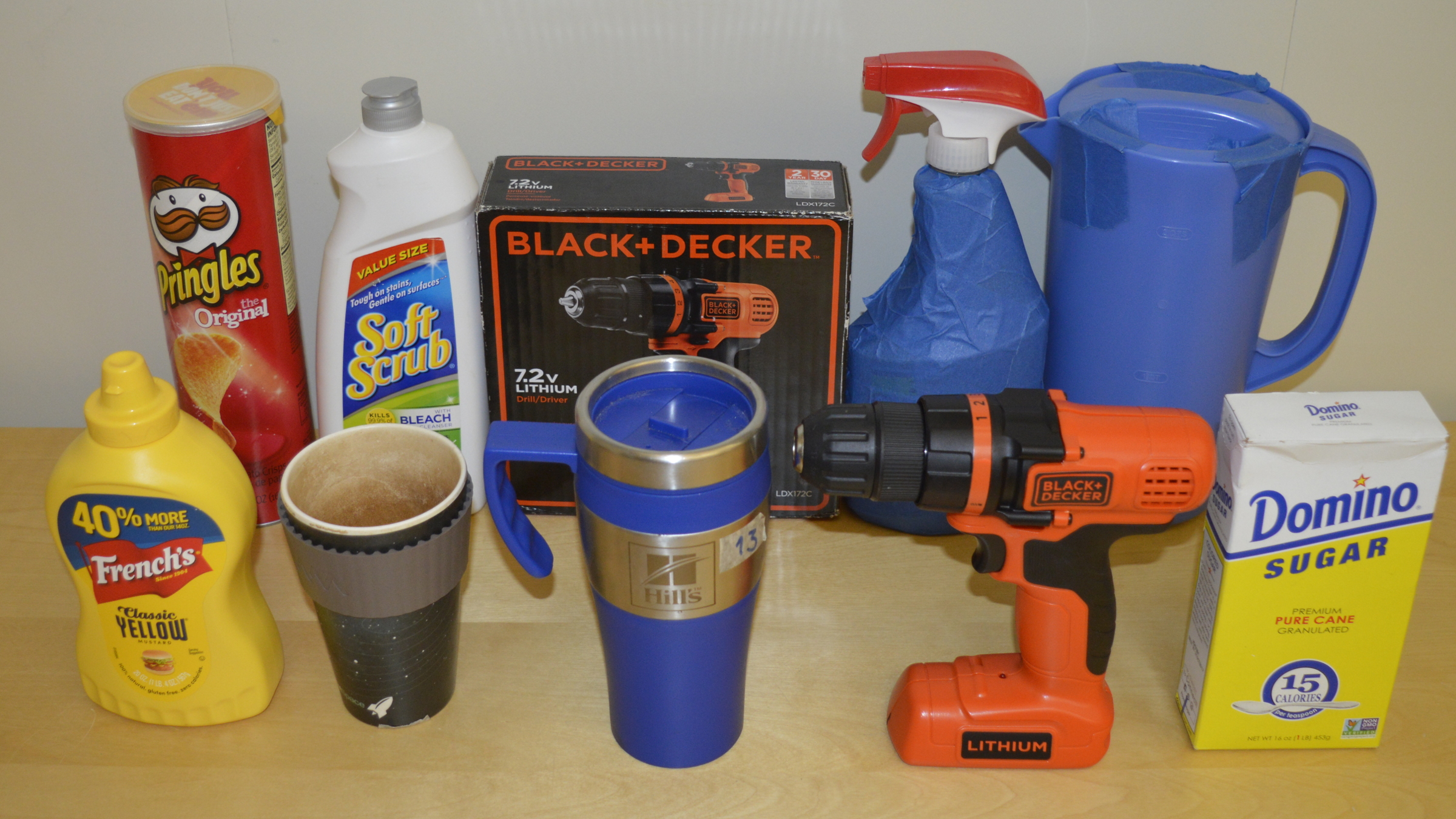}
	\caption{\label{fig:objects}The objects used in the physical experiments.}
 \figvspace{}
\end{figure}

\begin{figure*}\hspace{-0.5em}
	\centering
	\begin{subfigure}[b]{.1\textwidth}
		\centering
		\includegraphics[width=\linewidth]{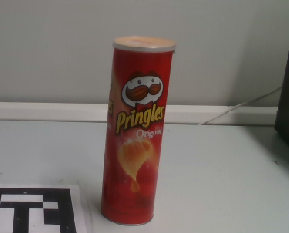}%
	\end{subfigure}\hspace{-0.3em}
	\begin{subfigure}[b]{.1\textwidth}
		\centering
		\includegraphics[width=\linewidth]{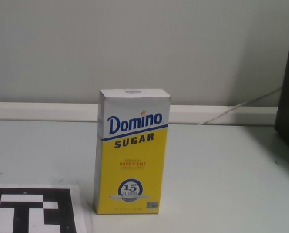}%
	\end{subfigure}\hspace{-0.3em}
	\begin{subfigure}[b]{.1\textwidth}
		\centering
		\includegraphics[width=\linewidth]{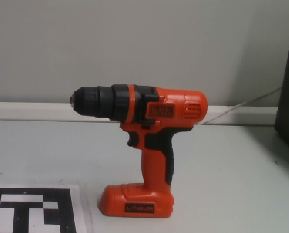}%
	\end{subfigure}\hspace{-0.3em}
	\begin{subfigure}[b]{.1\textwidth}
		\centering
		\includegraphics[width=\linewidth]{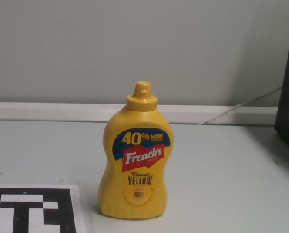}%
	\end{subfigure}\hspace{-0.3em}
	\begin{subfigure}[b]{.1\textwidth}
		\centering
		\includegraphics[width=\linewidth]{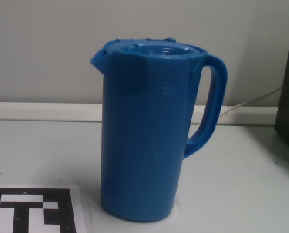}%
	\end{subfigure}\hspace{-0.3em}
	\begin{subfigure}[b]{.1\textwidth}
		\centering
		\includegraphics[width=\linewidth]{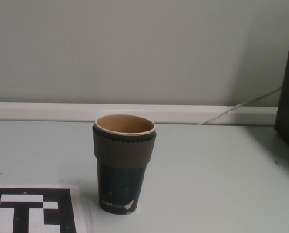}%
	\end{subfigure}\hspace{-0.3em}
	\begin{subfigure}[b]{.1\textwidth}
		\centering
		\includegraphics[width=\linewidth]{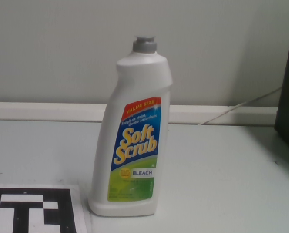}%
	\end{subfigure}\hspace{-0.3em}
	\begin{subfigure}[b]{.1\textwidth}
		\centering
		\includegraphics[width=\linewidth]{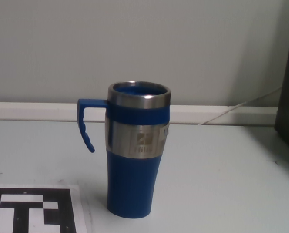}%
	\end{subfigure}\hspace{-0.3em}
	\begin{subfigure}[b]{.1\textwidth}
		\centering
		\includegraphics[width=\linewidth]{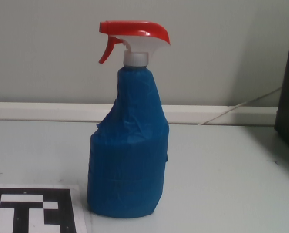}%
	\end{subfigure}\hspace{-0.3em}
	\begin{subfigure}[b]{.1\textwidth}
		\centering
		\includegraphics[width=\linewidth]{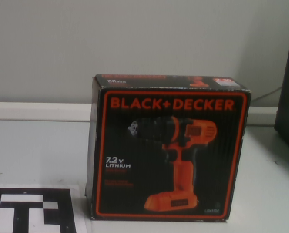}%
	\end{subfigure}
    \\\hspace{-0.5em}
	\begin{subfigure}[b]{.1\textwidth}
		\centering
		\includegraphics[width=\linewidth]{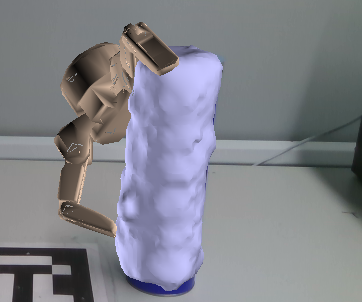}%
	\end{subfigure}\hspace{-0.3em}
	\begin{subfigure}[b]{.1\textwidth}
		\centering
		\includegraphics[width=\linewidth]{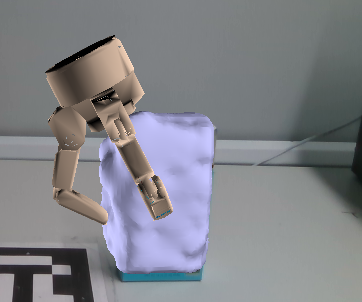}%
	\end{subfigure}\hspace{-0.3em}
	\begin{subfigure}[b]{.1\textwidth}
		\centering
		\includegraphics[width=\linewidth]{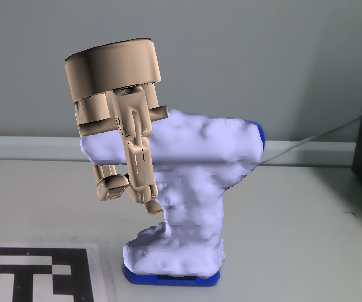}%
	\end{subfigure}\hspace{-0.3em}
	\begin{subfigure}[b]{.1\textwidth}
		\centering
		\includegraphics[width=\linewidth]{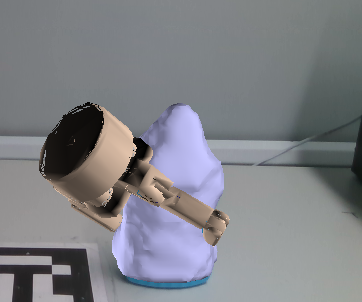}%
	\end{subfigure}\hspace{-0.3em}
	\begin{subfigure}[b]{.1\textwidth}
		\centering
		\includegraphics[width=\linewidth]{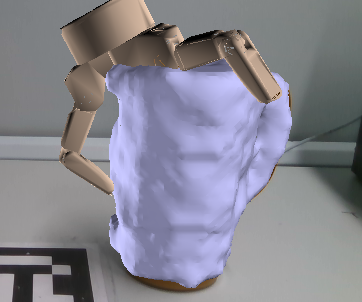}%
	\end{subfigure}\hspace{-0.3em}
	\begin{subfigure}[b]{.1\textwidth}
		\centering
		\includegraphics[width=\linewidth]{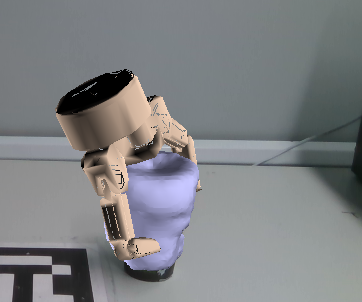}%
	\end{subfigure}\hspace{-0.3em}
	\begin{subfigure}[b]{.1\textwidth}
		\centering
		\includegraphics[width=\linewidth]{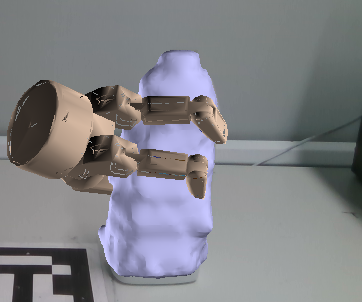}%
	\end{subfigure}\hspace{-0.3em}
	\begin{subfigure}[b]{.1\textwidth}
		\centering
		\includegraphics[width=\linewidth]{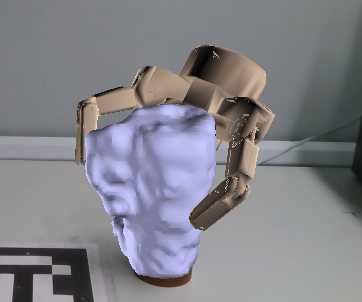}%
	\end{subfigure}\hspace{-0.3em}
	\begin{subfigure}[b]{.1\textwidth}
		\centering
		\includegraphics[width=\linewidth]{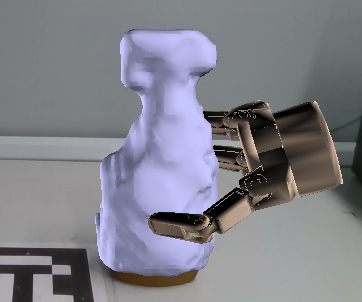}%
	\end{subfigure}\hspace{-0.3em}
	\begin{subfigure}[b]{.1\textwidth}
		\centering
		\begin{overpic}[width=\linewidth]{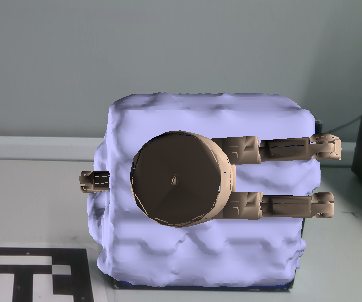}%
		\put(0,0){\tikz \draw[red,line width=0.5mm] (0,0) rectangle (1.75,1.45);}
		\end{overpic}
	\end{subfigure}
	\caption{\label{fig:grasps_on_real_obj}Example grasps proposed by \methodname{} on real objects. The upper row shows the unsegmented input image and the bottom rows shows some grasps on the shape-completed object. The grasp shown in the red box failed as it was in collision with the object.}
	 \figvspace{}
\end{figure*}
 
To further evaluate the impact  the  proposed loss functions have on performance, we conducted an ablation study where we trained models with one of the following losses removed: the interpenetration loss $\mathcal{L}_{int}$, the contact loss $\mathcal{L}_{cont}$, the orientation loss $\mathcal{L}_{orient}$, and the discriminator loss $\mathcal{L}_{disc}$. We train each of these models as was described in \secref{sec:exp_setup} keeping the same weights and evaluated them on the \egad{} dataset by sampling 110 grasps from 5 different viewpoints and calculating the $\epsilon$-quality and the intersection for the top 10 grasps.

The results are presented in Table \ref{tb:ablation_study}. As expected, a network trained with no interpenetration loss $\mathcal{L}_{int}$ often intersects the object as this results in more contacts but the final $\epsilon$-quality 0.77 is still not higher than 0.86 achieved with the model trained with all the losses. 

Another interesting observation is that the model with no contact loss $\mathcal{L}_{cont}$ still achieves a high $\epsilon$-quality. Our hypothesis was that grasps generated with this model would not interpenetrate the object at all as the model would have learned to translate the gripper far from the object which would also have resulted in low quality. However, this was not the case and one possible explanation why is that the orientation and discriminator losses forces the grasps to be realistic and oriented towards the object.    

A model trained without the orientation loss $\mathcal{L}_{orient}$ barely impacts the quality of the grasps but does increase the interpenetration compared to the full model. However, this loss speeds up learning in the early stages of training as it acts as an inductive bias forcing the hand to be oriented towards the object.

The network with no discriminator loss  $\mathcal{L}_{disc}$ produces the lowest quality grasps. At the same time, it also produces gripper joint-configurations that are physically infeasible. 

Overall, all of the models produce grasps with lower $\epsilon$-quality and higher interpenetration compared to a model trained with all the losses. However, the grasp quality does not heavily deteriorate. This is probably due to the inherent power of the finger refinement layer which will always refine the gripper's fingers close to the object if that is possible. Nevertheless, the ablation study shows that all the losses have different purposes and leaving one out affects the final performance of the model. Therefore, we use the model with all losses in our following Sim-to-Real experiment.

\subsection{Sim-to-Real Grasp Transfer}
\label{exp:sim_to_real}

\begin{table}
    \centering
	\ra{1.3}\tbs{10}
	\caption{\label{tb:real_exp_summary}Real hardware experiment results}
    \begin{tabular}{@{}lcc@{}}
        \toprule
        & \multicolumn{1}{l}{GraspIt!} & \multicolumn{1}{l}{\methodname{}} \\
        \midrule
        Grasp Success Rate (\%) $\uparrow$        & 40   & \textbf{60} \\
        Shape Completion Time (sec.) $\downarrow$ & 7.7  & \textbf{7.4} \\
        Grasp Evaluation Time (sec.) $\downarrow$ & 34.4 & \textbf{1.7} \\
        \bottomrule
    \end{tabular}
     \figvspace{}
\end{table}

To understand if grasps generated with our generative grasp sampler trained on synthetic data, transfer well to real objects, we conducted an experiment on a real \panda{} equipped with a \barrett{}. The goal was to grasp the objects shown in \figref{fig:objects} which were chosen as they represent a high variability in both size and shape.

To capture the RGB-D image we used an Intel RealSense D435 camera looking at the scene from the side at a 45 degree viewpoint. For the extrinsic calibration of the camera we used an Aruco marker~\cite{garrido2014automatic}. To provide our network with an RGB image of only the object, we segment it from the scene by subtracting the background and the table. To create a mesh of the segmented object, which is needed in both our method and the baseline, we used the shape-completion method detailed in~\cite{lundell2020beyond}. 
For both methods we generated 20 grasps per object. We then calculated the intersection and quality metric of each grasp. The first physically reachable grasp with lowest intersection and highest quality metric was executed on the real robot. To evaluate if a grasp was successful, the robot had to grasp the object and, without dropping it, move to the start position and rotate the hand $\pm$90° around the last joint. If the object was dropped during the manipulation, the grasp was considered unsuccessful.

The result of this experiment is shown in Table \ref{tb:real_exp_summary}. Based on these numbers we can see that our method reaches a grasp success rate of 60\% compared to the baseline 40\%, while being over 20 times faster. 

An example of the input image fed to the network and a generated grasp is shown in \figref{fig:real_experiment}. Although this image is not qualitatively as good as the training data visualized in \figref{fig:synthetic_training_data} the method was still able to generate high quality grasps on such objects showing stable sim-to-real transfer. Additional grasps generated using our method are visualized in \figref{fig:grasps_on_real_obj}. 

Based on the experiments, \methodname{} never produced grasps that were too far from the object to be able to grasp it. The main reason for grasp failure was that grasp ranking based on the $\epsilon$-quality metric favored grasps that established many contact points with the object and, as shown in the leftmost image in \figref{fig:grasps_on_real_obj}, such grasps may not translate to good real-world grasps. Another reason for grasp failure was object scale: on big objects the generated grasps were always in collision with the object, as shown in the rightmost image in \figref{fig:grasps_on_real_obj}. Despite these limitations, the results still indicate stable sim-to-real grasp transfer.

\section{Conclusions and future work}
\label{sec:conclusions}

We presented \methodname{}, a generative grasp sampling method that produces multi-fingered 6D grasps directly from an RGB-D image. The key insight was to reduce the search space by using a coarse-to-fine grasp generation method where we first generated coarse grasps based on a grasp taxonomy which subsequently were refined using a fully differentiable forward kinematics layer. We compared our model to the well known simulated annealing planner in \graspit{} both in simulation and on a real robot. The results showed that our model trained on synthetic data was significantly better than the baseline in generating higher quality grasps in simulation, and on real hardware it achieved a higher grasp success rate. At the same time it was also 20--30 times faster than the baseline. 

Despite the good results, there is still room for improvements. Although the classification into grasp types reduced the search space and eased the learning of the model it requires a large dataset of labeled grasps which is time-consuming to gather. A more elegant solution is to not classify grasps according to a taxonomy but instead regress directly to joint angles allowing to train the model on other datasets such as the Columbia grasp database~\cite{goldfeder2009columbia}. Another limitation is the computational time to evaluate the grasps which accounts for more than half the time needed to generate grasps. This time could be reduced by training a critic to evaluate multi-finger grasps but that is still an open problem.

In conclusion the work presented here shows that generating 6D coarse-to-fine multi-fingered grasps is both fast and leads to good grasps. This, in turn, opens the door to use dexterous hands for feedback-based grasping, task informative grasping and grasping in clutter.%
\newpage

\newpage

\bibliographystyle{IEEEtran}
\bibliography{refs}

\end{document}